\documentclass{article}

\usepackage{spconf,amsmath,graphicx}

\usepackage{enumitem}
\usepackage{mathtools,amssymb}
\usepackage{mwe} 
\usepackage{algorithmic}
\usepackage{textcomp}
\usepackage{array}
\usepackage{ragged2e}
\usepackage{multirow}
\usepackage{caption}
\usepackage{subcaption}
\usepackage{mathtools}
\usepackage{url}
\usepackage{float}

\usepackage{xcolor}
\definecolor{cb_orange}{rgb}{1.0,0.51,0.0}
\definecolor{cb_blue}{rgb}{0.22,0.49,0.72}
\definecolor{cb_green}{rgb}{0.3,0.67,0.29}
\definecolor{cb_red}{rgb}{0.89,0.1,0.11}
\definecolor{cb_purple}{rgb}{0.6, 0.31, 0.64}
\definecolor{cadetgrey}{rgb}{0.57, 0.64, 0.69}
\definecolor{cb_black}{rgb}{0, 0, 0}

\title{MED-TEX: Transfer and Explain Knowledge with Less Data from Pretrained Medical Imaging Models}
\name{Author(s) Name(s)}
\address{}
\name{Thanh Nguyen-Duc\sthanks{thanh.nguyen4@monash.edu, $^{\dagger}$ corresponding author, $^{\dagger}$Monash University}$^{\ddagger}$ \qquad He Zhao$^{\dagger \ddagger}$ \qquad Jianfei Cai$^{\ddagger}$ \qquad Dinh Phung$^{\ddagger}$}
\begin{document}
	\setlength{\belowdisplayskip}{0pt} \setlength{\belowdisplayshortskip}{0pt}
	\setlength{\abovedisplayskip}{0pt} \setlength{\abovedisplayshortskip}{0pt}
	
	%
	\maketitle
	\begin{abstract}
		\vspace{-0.15cm}
		Deep learning methods usually require a large amount of training data and lack interpretability.
		In this paper, we propose a novel knowledge distillation and model interpretation framework for medical image classification that jointly solves the above two issues. Specifically, to address the data-hungry issue, a small student model is learned with less data by distilling knowledge from a cumbersome pretrained teacher model.
		To interpret the teacher model and assist the learning of the student, an explainer module is introduced to highlight the regions of an input that are important for the predictions of the teacher model.
		Furthermore, the joint framework is trained by a principled way derived from the information-theoretic perspective.
		Our framework outperforms on the knowledge distillation and model interpretation tasks compared to state-of-the-art methods on a fundus dataset.
	\end{abstract}

	\vspace{-0.37cm}

\renewcommand{\vec}{\boldsymbol}
\newcommand{\pprob}{\operatorname{p}\probarg}
\DeclarePairedDelimiterX{\probarg}[1]{(}{)}{%
  \ifnum\currentgrouptype=16 \else\begingroup\fi
  \activatebar#1
  \ifnum\currentgrouptype=16 \else\endgroup\fi
}

\newcommand{\qprob}{\operatorname{q}\probargq}
\DeclarePairedDelimiterX{\probargq}[1]{(}{)}{%
  \ifnum\currentgrouptype=16 \else\begingroup\fi
  \activatebar#1
  \ifnum\currentgrouptype=16 \else\endgroup\fi
}
\newcommand{\innermid}{\nonscript\;\delimsize\vert\nonscript\;}
\newcommand{\activatebar}{%
  \begingroup\lccode`\~=`\|
  \lowercase{\endgroup\let~}\innermid
  \mathcode`|=\string"8000
}
\newcommand{\matr}[1]{\mathbf{#1}}
\newcommand{\diff}{\mbox{d}\,}

\newcommand{\matrl}[2]{\mathbf{#1}^{(#2)}}

\newcommand{\vecl}[2]{\vec{#1}^{(#2)}}

\newcommand{\kprime}{k^{\prime}}

\newcommand{\scall}[2]{#1^{(#2)}}

\newcommand{\bcdot}{\boldsymbol{\cdot}}

\newcommand{\expt}[1]{\mathbb{E}\left[#1\right]}
\newcommand{\exptt}[2]{\mathbb{E}_{#1}\left[#2\right]}
\newcommand{\eqdef}{\vcentcolon=}
\newcommand{\sigmoid}[1]{\text{sigmoid}\left(#1\right)}

\newcommand{\softmax}[1]{\text{softmax}\left(#1\right)}

\newcommand{\indicator}[1]{\textbf{1}_{\left[#1\right]}}
\newcommand{\rowa}[1]{\renewcommand{\arraystretch}{#1}}

\newcommand{\best}[1]{\textbf{#1}}
\newcommand{\second}[1]{\underline{#1}}

\newcommand{\ttt}{\mathcal{T}}
\newcommand{\sss}{\mathcal{S}}
\newcommand{\eee}{\mathcal{E}}
	\section{Introduction}
\begin{figure*}[t]
		\centering
		\includegraphics[width=0.85\linewidth]{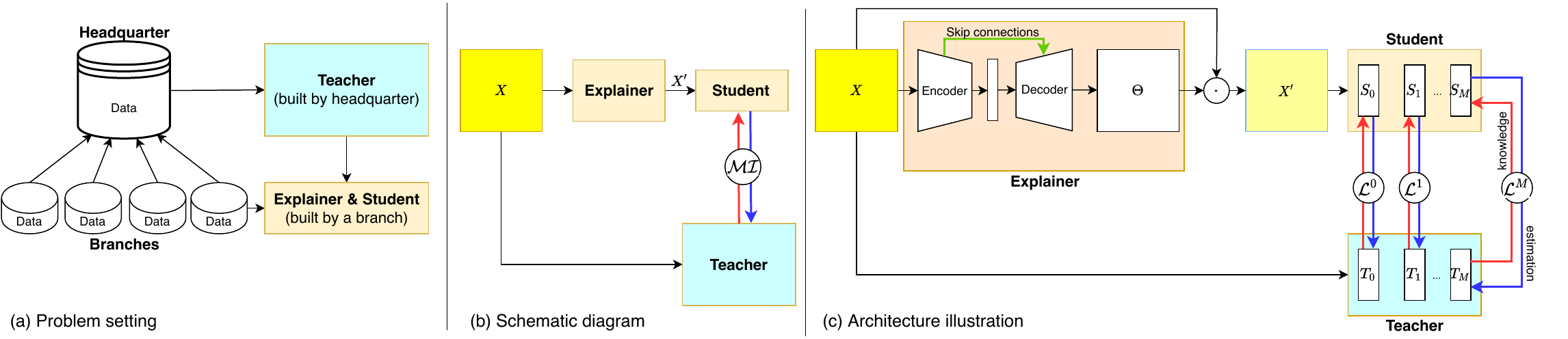}
		\caption{(a) Problem setting: a headquarter gathers data from multiple branches to produce a shared cumbersome teacher. A branch builds a local small and interpretable model. (b) An overview of our framework: fixed pretrained teacher, learnable explainer and learnable student. The explainer explains to the student by producing a simplified $\matr{X'}$ from input $\matr{X}$. The knowledge from teacher is transferred to the student by maximizing the mutual information ($MI$). (c) The detailed architecture.}
		\label{fig:overall}
		\vspace{-0.55cm}
	\end{figure*}

	\vspace{-0.25cm}
	A practical scenario of medical image classification applications~\cite{Rieke2020TheFO} is considered, where a central hospital headquarter gathers data from multiple local branches in Fig.~\ref{fig:overall}(a).
	The headquarter has developed a large CNN model for disease classification with excellent performance trained on a big dataset, which is the global model to be distributed to the branches.
	Given the limited computation, a branch wants to develop a customized smaller model using its local data. The branch cannot access to the big dataset of the headquarter because of privacy and sensitivity concerns. To assist the development of the local model, the knowledge from the global model is transferred to the local one~\cite{hinton2015distilling}. For medical domain, model interpretation is highly desirable. Therefore, the local model should have two capabilities: explaining the global model and transferring the knowledge of the global model to the local model with its local data only.\\
	\vspace{-0.03cm}
	\textit{Model perceptive interpretation} is defined by the ability to identify the areas of an input image that are important to the prediction of the classifier. Neural saliency such as Grad-CAM~\cite{selvaraju2017grad} is used to locate feature that contributes the most to the classification output. Feature selection~\cite{chen2018learning}, hard attention~\cite{jang2016categorical} and soft attention~\cite{xu2015show} are used to generate different weights for different features. However, they are not designed for explaining a pretrained global model. The recent Learning-to-Explain (L2X)~\cite{chen2018learning} trains an explainer to explain a pretrained global model by maximizing the mutual information between selected instance-wise features and the teacher outputs. L2X does not address the issue of lack of large training data and its effectiveness on high-resolutional image classification has not been confirmed. \\
	\vspace{-0.03cm}
	\textit{Knowledge distillation} is a process of transferring knowledge from the complicated global model (called teacher) to a smaller lighter-weighted one (called student).
	The small student model can significantly reduce the deployment cost of the local branch.
	KD was first introduced by Hinton et al.~\cite{hinton2015distilling} to distill knowledge from the distribution of class probabilities predicted by the teacher model. 
	Recently, Ahn et al.~\cite{ahn2019variational} exploited the information-theoretic perspective as maximizing the mutual information between the teacher and the student in order to transfer knowledge named (VID).
	In medical domain, Wang et al.~\cite{wang2019segmenting} used KD to train a student model that speeds up the inference time of a 3D neuron segmentation model.
	However, these previous approaches do not consider to interpreting the complicated teacher model.\\
	\vspace{-0.03cm}
	In this paper, we propose an end-to-end framework to address the above two requirements simultaneously by to learn a small medical image classification model with less training data but better interpretability.
	Our contributions follow: a) a new end-to-end MEDical Transfer and EXplain framework (MED-TEX) from a pretrained global model, which combines knowledge distillation and pixel-level model interpretation. Existing methods only focus on either of them; b) a joint training objective for our framework, derived from an information-theoretic perspective. It is both theoretically and practically appealing; c) experimental results demonstrate that our proposed method outperforms other methods on model interpretability and knowledge distillation.
	\vspace{-0.3cm}
	\vspace{-0.5cm}
	\section{Transfer and explain knowledge from medical pretrained models (MED-TEX)}
	\vspace{-0.2cm}
	\label{sec:method}

	We denote the CNN-based global classifier model (teacher, $\ttt$).
	The student $\sss$ is another CNN-based classifier that can potentially be a hundred times smaller to significantly reduce computational complexity and be trained by less data only from the local branch.
	With input image, $\matr{X} \in \mathbb{R}^{C \times H \times W}$ ($C, H, W$ are the channels, height and width of the image, respectively), the explainer $\eee$ inspired by under-completed auto encoder with skip connections produces the selection scores $\matr{\Theta}$, which are high for the important pixels for the decision of the teacher and low for the unimportant ones.
	In our framework, $\matr{\Theta}$ has the same size to $\matr{X}$ (at pixel level) and is element-wise multiplied by $\matr{X}$ to get a simplified the input image, denoted by $\matr{X}'$.
	This $\matr{X}'$ is then input to the student $\sss$ to perform predictions. Our goal is training the student to mimic the behaviors of the teacher by pushing teacher's outputs from the last and intermediate layers close to student's outputs while the explainer produces $\matr{\Theta}$ to guide the student by highlighting the regions of $\matr{X}$ to generate $\matr{X'}$, as illustrated in Fig.~\ref{fig:overall}(c).\\
	\vspace{-0.05cm}
	\textbf{Proposed framework.}
	The teacher's and student's predicted distributions over the labels is denoted $\vec{y}^{\ttt} \in \Delta^{L}$ and $\vec{y}^{\sss} \in \Delta^{L}$ respectively, 
	where $L$ is the number of labels and $\Delta^L$ denotes the $L$ dimensional simplex. 
	The $\ttt$ and $\sss$ has $M$ and $N$ layers respectively, where the last layer is a fully connected layer and other layers are convolutional layers (or block convolution layers). $N$ can be different from $M$ as in \cite{ahn2019variational}; however, we simplify formulas by $N=M$. 
	We have $\vec{y}^\ttt = \ttt(\matr{X})$ (i.e., $\pprob{y^\ttt_l | \matr{X}} \propto \ttt(\matr{X})_l$), $\matr{X}' | \matr{X} = \eee(\matr{X})$, and $\vec{y}^\sss | \matr{X}' = \sss(\matr{X}')$ (i.e.,  $\qprob{y^\sss_l | \matr{X'}} \propto \sss(\matr{X})_l$).
	We formulate our preliminary goals of explaining and extracting the teacher's knowledge to the student as the following loss derived from mutual information (derived from Eq.~(\ref{eq:term1_v})).
	%
	\begin{equation}
		\label{eq:aim1}
		\begin{split}
			\mathcal{L}^N = \min_{\eee, \sss} - \mathbb{E}_{\matr{X}} \Big[ \mathbb{E}_{\matr{X'} |  \matr{X} } \Big[ \mathbb{E}_{\vec{y}^\ttt |  \matr{X'}}[\log q(\vec{y}^\ttt | \matr{X'})] \Big] \Big],
		\end{split}
	\end{equation}
	where $q$ corresponds to our student, acting as the variational distribution in the deviation of mutual information. 
	Eq. (\ref{eq:aim1}) is similar to minimizing the cross-entropy loss between the outputs of the teacher and the student and generate $\matr{X'}$ by element-wise multiplication between $\matr{X}$ and $\matr{\Theta}$,
	aiming to push the predictions of the student close to those of the teacher, with the help from the explainer:
	\begin{equation}
		\begin{split}
			\label{eq:loss1}
			\mathcal{L}^N = \min_{\eee, \sss} -\exptt{\matr{X}}{\exptt{\matr{X}' | \matr{X}}{\sum_l^L \pprob{y^\ttt_l | \matr{X}} \log \qprob{y^\sss_l | \matr{X'}}}}.
		\end{split}
	\end{equation}

	Given an input image $\matr{X}$, the explainer generates an importance score for each of its pixels, where its last layer is $1\times1$ convolution layer with sigmoid activation.
	The higher the important score is, the more important the corresponding pixel is to the prediction of the teacher.
	All the importance scores form the importance map, denoted as $\matr{\Theta} \in [0,1]^{C \times H \times W}$. The output of the explainer can be expressed as:
	\begin{equation}
		\begin{split}
			\label{eq:explainer}
			\matr{X'} = \matr{\Theta}  \odot \matr{X},
		\end{split}
	\end{equation}
	where $\odot$ is the element-wise multiplication.
	
	
	Inspired by the idea of knowledge distillation in \cite{ahn2019variational}, we therefore introduce an additional loss to maximize the mutual information between the outputs of each $i^{\text{th}}$ intermediate layer of the teacher ($\ttt^i(\matr{X})$) and the student ($\sss^i(\matr{X'})$).  We simplify $i=j$ in $\ttt^i(\matr{X})$ and $\sss^j(\matr{X'})$ but it can be $i \neq j$ as in \cite{ahn2019variational}.
	\begin{equation}
		\begin{split}
			\label{eq:loss2}
			\mathcal{L}^{i} = \min_{\eee, \sss} -\exptt{\matr{X}}{\exptt{\matr{X}' | \matr{X}}{\log r (\ttt^{i}(\matr{X}) | \sss^{i}(\matr{X'}))  }},
		\end{split}
	\end{equation}
	where $\ r (\ttt^{i}(\matr{X}) | \sss^{i}(\matr{X'}))$ is a variational distribution used for approximating $\ \pprob{\ttt^{i}(\matr{X})|\sss^{i}(\matr{X'})}$, which is derived from information-theoretic perspective (see Eq.~\eqref{eq:term2_v}).
	
	Recall that the output of the $i^{\text{th}}$ layer of the teacher is a $C^i \times H^i \times W^i$ feature map (note that the output of the $i^{\text{th}}$ layer of the student is of the same spatial dimension but with a smaller number of channels). Following \cite{ahn2019variational}, we model $\ttt^i(\matr{X})$ as the following Gaussian distribution conditioned on $\sss^i(\matr{X'})$:
	\vspace{-0.25cm}
	\begin{equation}
		\begin{split}
			\label{eq:vkd:q}
			r\left(\ttt^i(\matr{X}) | \sss^i(\matr{X'})\right) \sim \prod_{c=1, h=1, w=1}^{C^i, H^i, W^i} \mathcal{N}\left(\mu^i(\sss^i(\matr{X'}))_{c, h, w}, \sigma^{i^2}_c\right), 
		\end{split}
	\end{equation}
	where $\mu^i$ is a subnetwork with $1 \times 1$ convolutional layers to match the channel dimensions between $\ttt^i(\matr{X})$ and $\sss^i(\matr{X'})$, $\mu^i_{c, h, w}$ is a single output unit, and $\sigma^{i^2}_c$ is the learnable parameter specific to each channel at the $i^{\text{th}}$ layer.
	For $\sigma^{i^2}_c$, we exploit the softplus function $\sigma^{i^2}_c = \log (1+e^{\alpha_c^{i}}) + \epsilon$  where $\alpha_c^i$ is a learnable parameter and $\epsilon$ is used for numerical stability. 
	
	With Eq.~(\ref{eq:vkd:q}), we can write Eq.~(\ref{eq:loss2}) as:
	\begin{equation}
		\begin{split}
			\label{eq:loss22}
			\mathcal{L}^{i} = \min_{\eee, \sss} \mathbb{E}_{\matr{X}}\Bigg[ \mathbb{E}_{\matr{X}' | \matr{X}}\Bigg[\sum_{c=1,h=1,w=1}^{C^i, H^i, W^i} \log \sigma^{i}_c + \\
			\frac{(\ttt^i(\matr{X})_{c,h,w} - \mu^i(\sss^i(\matr{X'}))_{c, h, w})^2}{2\sigma^{i^2}_c} + \text{const.}\Bigg]\Bigg].
		\end{split}
	\end{equation}
	
	Finally, the overall loss function of our framework can be written as
	\vspace{-0.25cm}
	\begin{equation}
		\begin{split}
			\label{eq:loss3}
			\mathcal{L} = \mathcal{L}^N + \lambda \sum_{i=1}^{N-1} \mathcal{L}^i,
		\end{split}
	\end{equation}
	where $\lambda$ is the weight of the losses of the intermediate layers.\\
	\textbf{Derivation from information-theoretic perspective.}
	\label{sec:derivation}
	Previously, the objective function of our proposed framework has intuitive interpretations.
	Here we additionally demonstrate that the objective function can be derived in a theoretical way with mutual information, which is a widely used measure of the dependence between two random variables and captures how much knowledge of one random variable reduces the uncertainty about the other.
	In particular, we note: minimizing the training losses in Eq.~(\ref{eq:loss1}) and Eq.~(\ref{eq:loss2}) are equal to maximizing the following mutual information: $I(\matr{X'}; \vec{y}^\ttt)$ and $I(\ttt^i(\matr{X}); \sss^i(\matr{X'}))$, respectively.
	%
	\begin{equation}
		\begin{split}
			\label{eq:objective}
			\max_{\eee, \sss} ~ I(\matr{X'}; \vec{y}^\ttt) +  \lambda  \sum_{i=1}^{N-1} I (\ttt^i(\matr{X}); \sss^i(\matr{X'})).
		\end{split}
	\end{equation}
	
	Given the definition of mutual information, the first term of Eq.~(\ref{eq:objective}) can be derived as:
	\begin{equation}
		\label{eq:term1}
		\begin{split}
			I(\matr{X'}; \vec{y}^\ttt) &= \mathbb{H}( \vec{y}^\ttt) -  \mathbb{H}( \vec{y}^\ttt| \matr{X'}) \\
			&= \mathbb{E}_{\matr{X}}\mathbb{E}_{\matr{\matr{X'}} |  \matr{X} }\mathbb{E}_{\vec{y}^\ttt |  \matr{X'}}[\log p(\vec{y}^\ttt | \matr{X'})] + Const.
		\end{split}
	\end{equation}
	In general, it is impossible to compute expectations under the conditional distribution of $p(\vec{y}^\ttt | \matr{X'})$.  
	%
	Hence, we define a variational distribution $q( \vec{y}^\ttt | \matr{X'} )$ that approximates $p( \vec{y}^\ttt | \matr{X'} )$:
	\begin{equation}
		\label{eq:term1_v}
		\begin{split}
			&\mathbb{E}_{\vec{y}^\ttt |  \matr{X'}}[\log p(\vec{y}^\ttt | \matr{X'})]
			= \mathbb{E}_{\vec{y}^\ttt |  \matr{X'}}[\log q(\vec{y}^\ttt | \matr{X'})]  \\
			&+ \mathbb{D}_{KL} [ q( \vec{y}^\ttt | \matr{X'} ) ||  p( \vec{y}^\ttt | \matr{X'}) ] 
			\geq  \mathbb{E}_{\vec{y}^\ttt |  \matr{X'}}[\log q(\vec{y}^\ttt | \matr{X'})],  \\
		\end{split}
	\end{equation}
	where $\mathbb{D}_{KL}$ is the Kullback–Leibler divergence and equality holds if and only if $q( \vec{y}^\ttt | \matr{X'} )$ and $p( \vec{y}^\ttt | \matr{X'} )$ are equal in distribution.
	Note that it is not hard to show that our student corresponds to the variational distribution $q$.
	
	For the second term of Eq.~(\ref{eq:objective}),
	we have:
	%
	\begin{equation}
		\label{eq:term2}
		\begin{split}
			I(\ttt&^i(\matr{X}); \sss^i(\matr{X'})) =  \mathbb{H}(\ttt^i(\matr{X})) - \mathbb{H} ( \ttt^i(\matr{X}) | \sss^i(\matr{X'})) \\
			= &\mathbb{E}_{\matr{X}}  \mathbb{E}_{\matr{\matr{X'}}  | \matr{X}} \mathbb{E}_{  \ttt^i| \matr{X} ,\sss^i | \matr{X'}}[\log p( \ttt^i(\matr{X}) | \sss^i(\matr{X'}))] + Const
		\end{split}
	\end{equation}
	Given Eq.~(\ref{eq:term2}), we can derive the following formula, similar to Eq.~(\ref{eq:term1_v}):
	\begin{equation}
		\label{eq:term2_v}
		\begin{split}
			\mathbb{E}_{  \ttt^i | \matr{X},\sss^i | \matr{X'}}&[\log p( \ttt^i(\matr{X}) | \sss^i(\matr{X'}))] \\
			\geq &   \mathbb{E}_{  \ttt^i | \matr{X},\sss^i | \matr{X'}}[\log r ( \ttt^i(\matr{X}) | \sss^i(\matr{X'}))], \\
		\end{split}
	\end{equation}
	where $r$ is the variational distribution to approximate the conditional distribution.
	
	By using the two variational distributions $q$ and $r$, the problem \eqref{eq:objective} can be relaxed to Eq.~\eqref{eq:ob_v}, i.e. maximizing the variational lower bounds.
	%
	\begin{equation}
		\begin{split}
			\label{eq:ob_v}
			\max_{\eee, \sss} ~ \mathbb{E}[\log q(\vec{y}^\ttt | \matr{X'})]  +  \lambda  \sum_{i=1}^{N-1} \mathbb{E}[\log r ( \ttt^i(\matr{X}) | \sss^i(\matr{X'}))].
		\end{split}
	\end{equation}
	
	\vspace{-0.8cm}
	\section{Experiments}
	\vspace{-0.2cm}
	\label{sec:result}
	In this section, we present the experiments conducted on a real-world dataset to evaluate the performance of the proposed MED-TEX against the state-of-the-art methods.\\
	\textbf{Architectures and settings of MED-TEX.}
	For the teacher and student, we adopt a deep architecture with 4 block CNN layers, where each block consists of a convolutional layer, batch normalization, maxpooling and ReLU activation.
	Due to the smaller number of filters, the size of the student model is much (226 times) smaller than the teacher, i.e., 1.7k parameters of the student versus 390.5k parameters of the teacher. We empirically figure out that the current teacher architecture works well with our fundus dataset. However, it is important to note that our framework is general enough to be applied to various teacher and student architectures. For the explainer, we adopt auto encoder with skip connections. The last layer of explainer is $1\times1$ convolution with sigmoid activation.\\
	\textbf{Dataset.} We conducted our experiment on a fundus dataset\footnote{Cao Thang Eye Hospital + https://ichallenge.baidu.com} with normal or abnormal\footnote{Myopia is an eye disease that causes distant objects to be blurry.} class.
	Finally, we have 1873 images in total, which consists of 1073 normal and 800 abnormal images. For the abnormal images, there are 200 of them with fine-grained lesion segmentation.
	The dataset is split into the training (773 normal and 500 abnormal images) and testing (300 normal and 300 abnormal images) sets. All the 200 images with lesion segmentation are in the testing set.
	We simulate less data scenario to learn by reducing the number of training images, i.e.,  25\% and 50\% training images are used, denoted as Fundus-25\% and Fundus-50\%, respectively.\\
	\textbf{Compared methods.}
	To our knowledge, there is no existing method that solves the exact same problem as ours. Thus, we individually compare our MED-TEX to knowledge distillation methods (e.g., KD\cite{hinton2015distilling} and VID\cite{ahn2019variational}) and model interpretation methods (e.g., hard attention using Gumbel-softmax trick\cite{jang2016categorical}, soft attention\cite{ xu2015show}, Grad-CAM\cite{selvaraju2017grad} and L2X\cite{chen2018learning}). 
	These comparison methods use ResNet18 backbone, which has significantly more parameters than the combination of explainer and student.
	To evaluate effectiveness of intermediate layer losses, we compare MED-TEX with its variant without information transfer losses (Eq.~\ref{eq:loss2}), denoted as MED-EX. 
	The importance of explainer is illustrated by using another variant, i.e., the student (only) without the explainer and intermediate  layer losses, which is only trained on the input image $X$. 
	All models are trained by using Adam with learning rate $0.001$,  $\lambda=0.01$ and batch size of 64.\\
	\textbf{Evaluation metrics.}
	Two metrics are introduced to be used to evaluate our framework.
	\textit{Post-hoc metric}~\cite{chen2018learning} compares the predictive distributions of the student given $\matr{X'}$ and the teacher given $\matr{X}$.
	In other words, we compute accuracy and f1 score by comparing between $\vec{y}^{\ttt}$ and $\vec{y}^{\sss}$ for knowledge distillation evaluation. Note that these post-hoc metrics do not compared to human labels.\\
	\textit{Intersection over Union (IoU)} compares between the highlighted image regions and the ground-truth lesion segmentation of abnormal images for interpretation evaluation. For a better comparison, we rank feature scores and select the number of pixels corresponding to the top $K$ highest scores (e.g., top$K$ $\in \{k\times 32 \times 32 ~ | ~ k = 1, 2, 3, 4, 5, 6\}$):
	\begin{equation}
		\begin{split}
			\label{eq:iou}
			IoU_{topK} = 2 \frac{\matr{\Theta}_{topK} \bigcap \matr{X}_{lesion}}{\matr{\Theta}_{topK} \bigcup \matr{X}_{lesion}},
		\end{split}
	\end{equation}
	where $\matr{\Theta}_{topK}$ indicates the selected pixels corresponding to the top$K$ feature scores 
	and $\matr{X}_{lesion}$ denotes ground-truth lesion segmentation pixels.\\
	\begin{table}[t]
		\centering
		\caption{Post-hoc evaluation on the fundus dataset.}
		\label{table:posthoc}
		\begin{tabular}{ |c||c|c||c|c||}
			\hline
			& \multicolumn{2}{c||}{Fundus-25\%} & \multicolumn{2}{c||}{Fundus-50\%} \\
			\hline
			Method & Acc & F1 & Acc & F1\\
			\hline 
			ResNet18+KD \cite{hinton2015distilling} & 0.863 & 0.901 & 0.931 & 0.945 \\
			\hline 
			ResNet18+VID \cite{ahn2019variational} & 0.891 & 0.911 & 0.921 & 0.937 \\
			\hline \hline
			Student (only) &  0.863 & 0.856 & 0.90 & 0.897  \\  
			\hline 
			MED-EX  & \underline{0.908} & \underline{0.925} & \underline{0.938} & \underline{0.950}  \\ 
			\hline 		
			MED-TEX &  \textbf{0.915} & \textbf{0.933} & \textbf{0.955} &  \textbf{0.964} \\
			\hline
		\end{tabular}
	\end{table}
	\vspace{-0.5cm}
	\begin{figure}[t]
		\centering	
		
		\begin{subfigure}{.49\linewidth}
			\centering
			\includegraphics[width=1\linewidth]{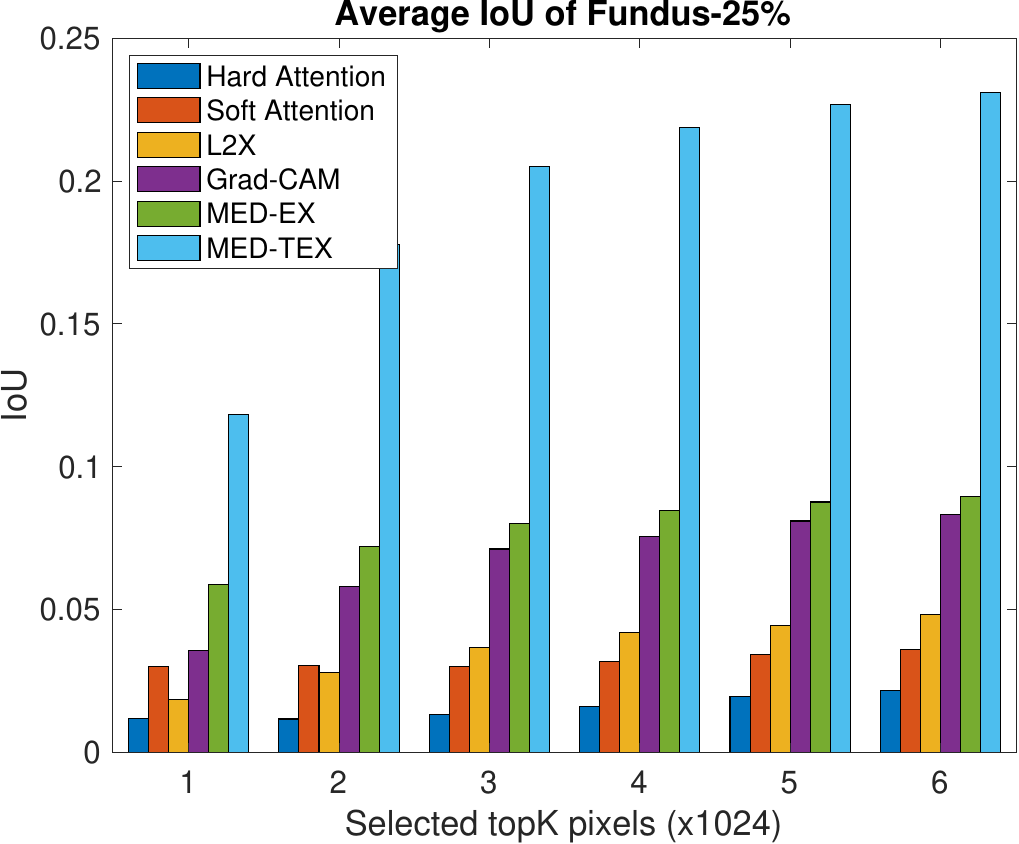}
			\caption{Fundus-25\%}\label{fig:image21}
		\end{subfigure}
		\hfill
		\begin{subfigure}{.49\linewidth}
			\centering
			\includegraphics[width=1\linewidth]{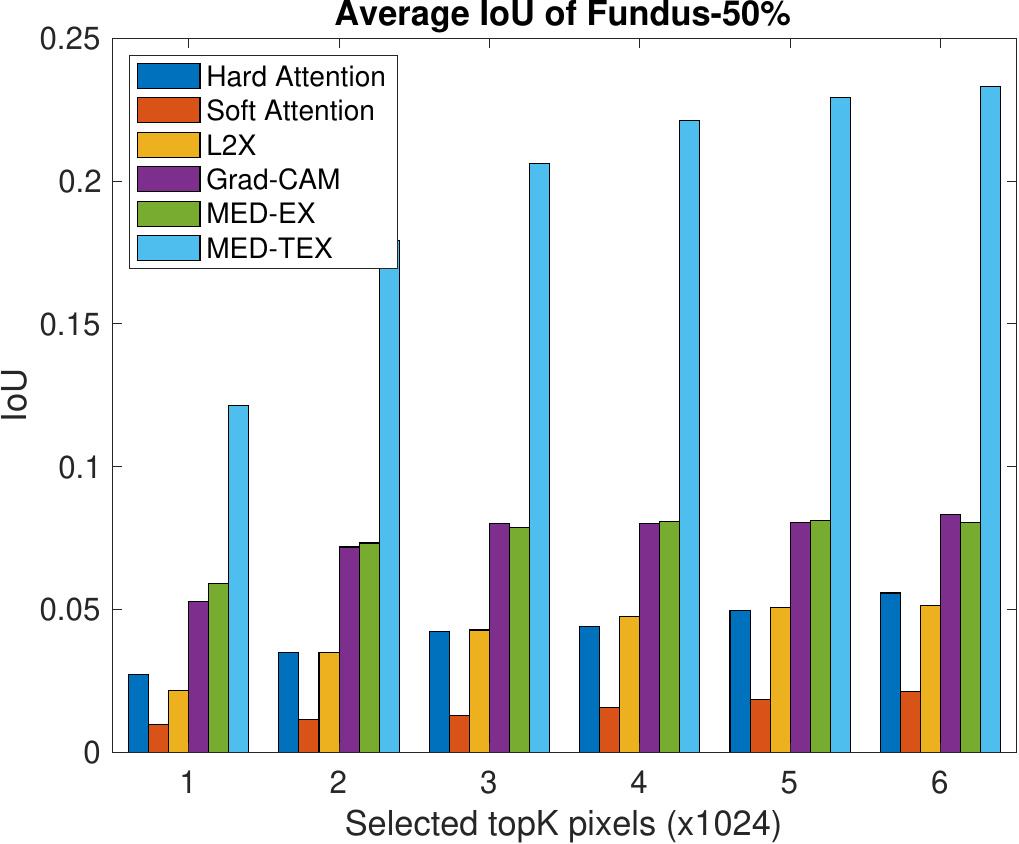}
			\caption{Fundus-50\%}\label{fig:image22}
		\end{subfigure}
		
		\caption{Average IoU of compared methods at different topKs.}
		\label{fig:iou}
	\end{figure}
	
	
	\begin{figure}[t]
		\centering
		\includegraphics[width=0.63\linewidth]{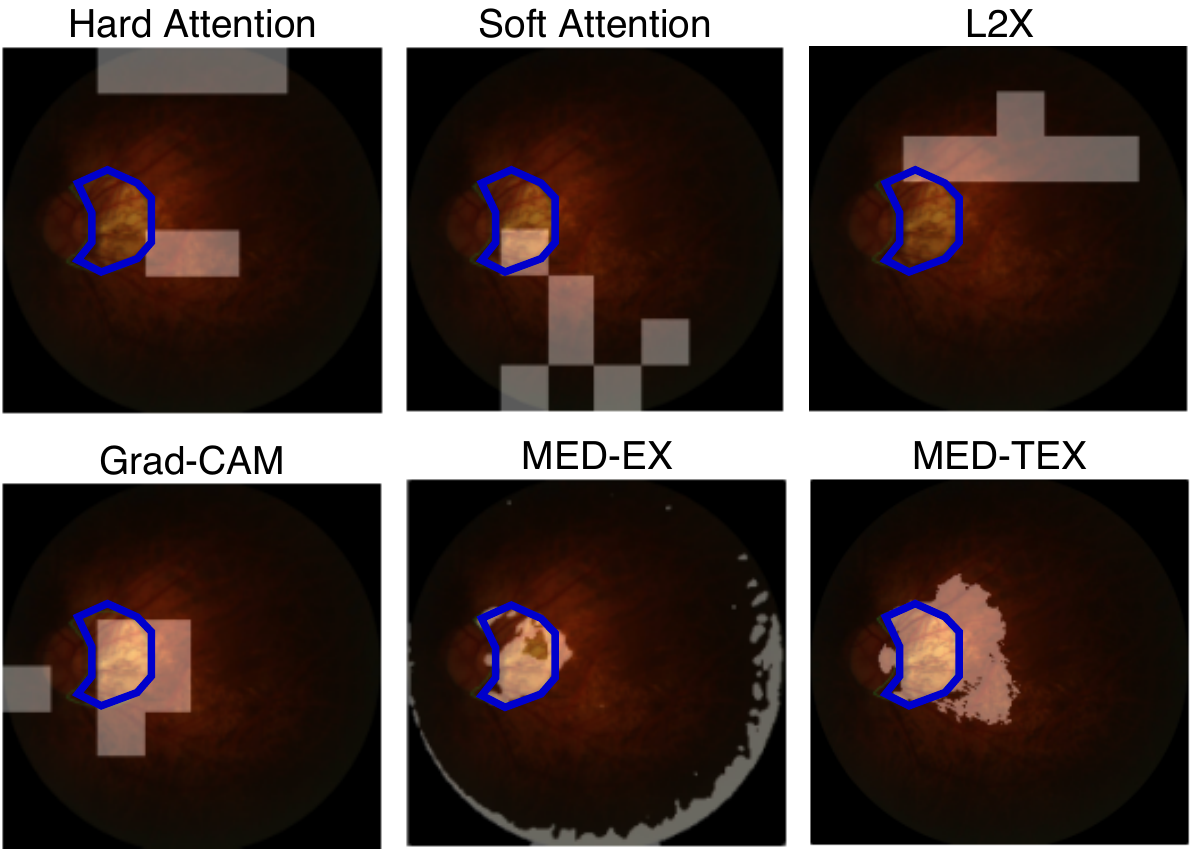}
		\caption{Methods evaluation of highlighted image region outputs (white) and the contour ground-truth lesion (blue).
		}
		\label{fig:fundus}
	\end{figure}
	
	\begin{figure}[t]
		\centering
		\includegraphics[width=0.65\linewidth]{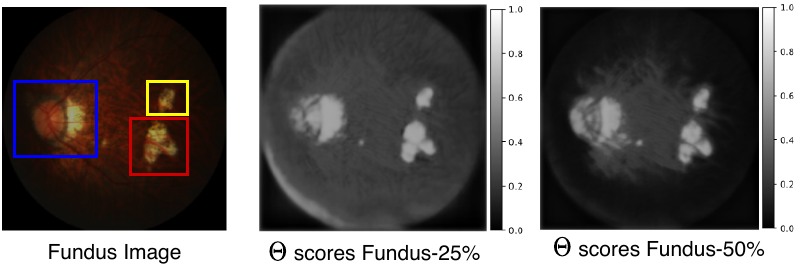}
		\caption{Visual feature selection scores $\Theta$ compared to three lesions (blue, yellow, red bounding boxes).}
		\label{fig:fundusvkd}
		\vspace{-0.3cm}
	\end{figure}
	\textbf{Results.} 
	\textit{For knowledge distillation} performance shown in Table~\ref{table:posthoc}, in terms of the post-hoc metric~\cite{chen2018learning},
	our method consistently outperforms other methods in term of both accuracy and F1 score.
	The Student (only) trained directly from raw input images cannot perform well. This suggests that the explainer with feature selection at pixel-level plays a critical role to guide the student to achieve better performance.
	The MED-TEX outperforms MED-EX, which indicates that it is beneficial to leverage the information in the intermediate layers.\\
	\textit{For model interpretation}, Fig.~\ref{fig:iou}  shows the IoU results in bar charts. 
	Our explainer of MED-TEX achieves significantly higher IoU than others.
	Fig.~\ref{fig:fundus} shows the visualization results of top$K$=6 highlighted image regions of different methods. 
	Hard attention, soft attention, Grad-CAM and L2X can only give patch-based region selection maps, while our MED-EX and MED-TEX produces pixel-level selection scores.
	Our method highlights the lesion regions that well match the ground-truth lesion in  Fig.\ref{fig:fundus}.
	Moreover, MED-TEX clearly outperforms MED-EX because of the intermediate knowledge distillation losses as shown in Fig.~\ref{fig:fundusvkd}.
	\vspace{-0.6cm}
	\section{Conclusion}
	\vspace{-0.4cm}
	\label{sec:conclusion}
	In this paper, we have introduced our novel framework MED-TEX, which is a joint knowledge distillation and model interpretation framework that learns the significantly smaller student (compared to the teacher) and explainer models by leveraging the knowledge only from the pretrained teacher model.
	In our experiment, we show that MED-TEX outperforms several widely used knowledge distillation and model interpretation techniques.
	\vspace{-0.4cm}
	\bibliographystyle{IEEEbib}
	\bibliography{refs}
	 \begin{figure*}[b]
	 	\centering
	 	\includegraphics[width=0.8\linewidth]{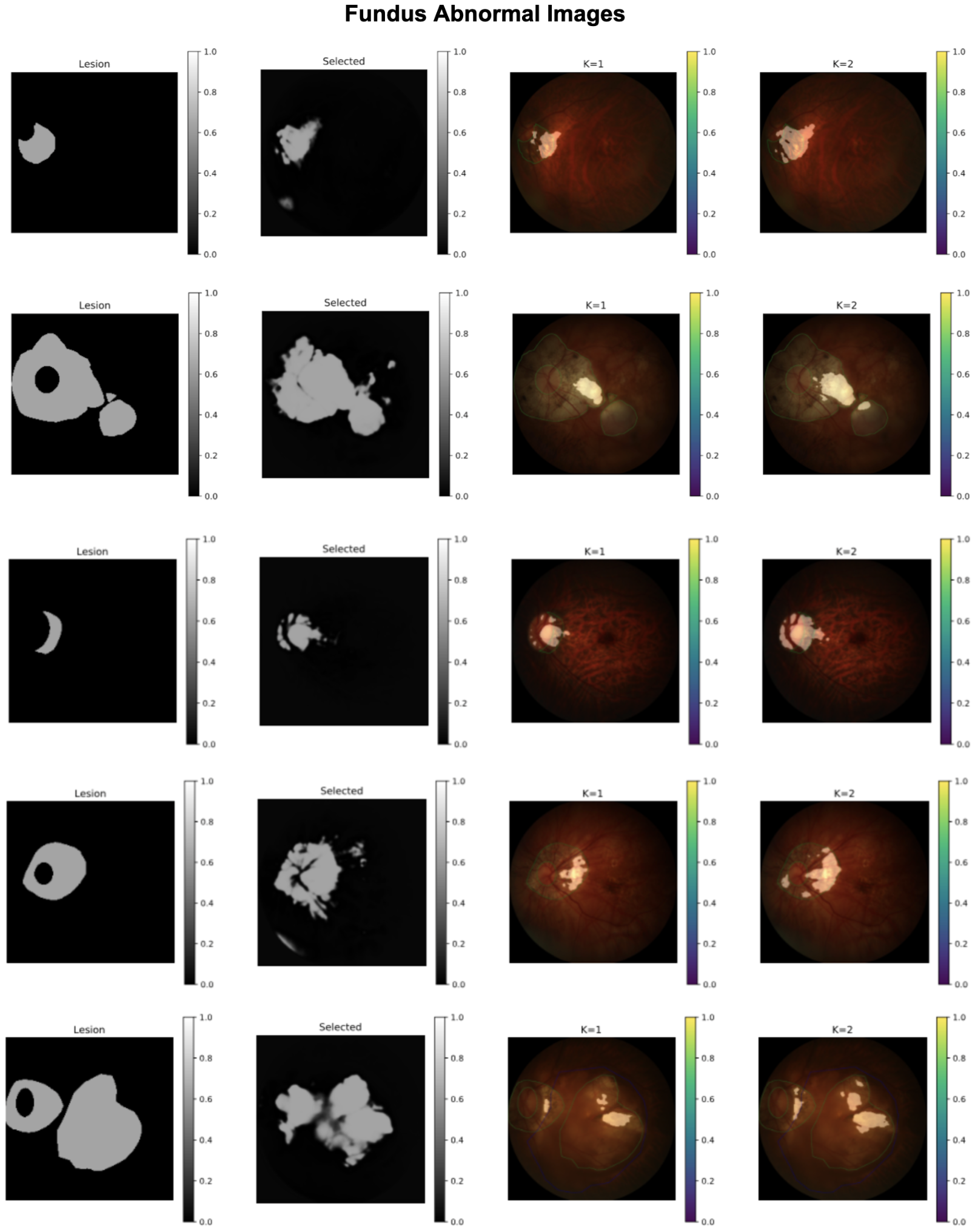}
	 	\caption{Additional Visual Results of MED-TEX: (Lesion, Selected and K indicate for expert segmentations, feature selection scores and topK x1024, respectively).}
	 \end{figure*}
\end{document}